\pdfoutput=1

\documentclass[11pt]{article}

\usepackage[]{naacl2021}

\usepackage{times}
\usepackage{latexsym}
\usepackage{enumitem}

\usepackage[T1]{fontenc}

\usepackage[utf8]{inputenc}
\usepackage{graphicx} 
\usepackage[]{algorithm2e}
\usepackage{pbox}
\usepackage{multirow}%
\usepackage{graphicx}
\usepackage{caption,subcaption}
\usepackage{hhline}
\usepackage{amsmath,amssymb}
\usepackage{booktabs}
\usepackage{microtype}

\title{Extractive Research Slide Generation Using Windowed Labeling Ranking}

\author{Athar Sefid \\
  Pennsylvania State University  \\
  \texttt{atharsefid@gmail.com} \\\And
  
  Jian Wu\\
  Old Dominion University  \\
  \texttt{j1wu@odu.edu} \\\AND
  
  Prasenjit Mitra\\
  Pennsylvania State University  \\
  \texttt{pum10@psu.edu} \\\And
  
  C Lee Giles\\
  Pennsylvania State University  \\
  \texttt{clg20@psu.edu} \\
  }

\begin{document}
\maketitle

\begin{abstract}
Presentation slides describing the content of scientific and technical papers are an efficient and effective way to present that work. However, manually generating presentation slides is labor intensive. We propose a method to automatically generate slides for scientific papers based on a corpus of 5000 paper-slide pairs compiled from conference proceedings websites. The sentence labeling module of our method is based on SummaRuNNer, a neural sequence model for extractive summarization. Instead of ranking sentences based on semantic similarities in the whole document, our algorithm measures importance and novelty of sentences by combining semantic and lexical features within a sentence window. Our method outperforms several baseline methods including SummaRuNNer by a significant margin in terms of ROUGE score. The code and data is available \href{https://github.com/atharsefid/Extractive_Research_Slide_Generation_Using_Windowed_Labeling_Ranking}{here} .
\end{abstract}

\section{Introduction}

It has become common practice for researchers to use slides as a visual aid in presenting research findings and innovations. 
Such slides usually contain bullet points that the researchers believe to be important to show. These bullet points serve both as a reminder to the speaker (when he/she is presenting) and summaries for audiences to understand.  
Manually creating a set of high-quality slides from an academic paper is time-consuming. 
We propose a method that automatically selects salient sentences that could be included into the slides, with the purpose of reducing the time and effort for slide generation. 

The main challenge for solving this problem is to accurately extract the main points from an academic paper. This is due to the limitations of existing methods to fully encode semantics of sentences and the implicit relations between sentences. 
Here, we propose an extractive summarizer that identifies the best sentence in a set of consecutive sentence windows. The selection process depends on importance and novelty of the sentence that is modeled by the neural networks. The selected sentences and their frequent noun phrases are then structured in a layered format to make the bullet points of the slides. 

Presentation slides are usually created with multiple bullet points organized in a multi-level hierarchical structure, usually with phrases summarizing high level topics at the first level and bullets at the second and other levels for further clarification or details. Statistical analysis on our training data set shows that more than 92\% of the bullets are in the first and second level and only 8\% are in the third layer. Therefore, we built our presentations in two level bullet points only.
\begin{figure}[tbp]
\includegraphics[width=0.48\textwidth, height=7cm]{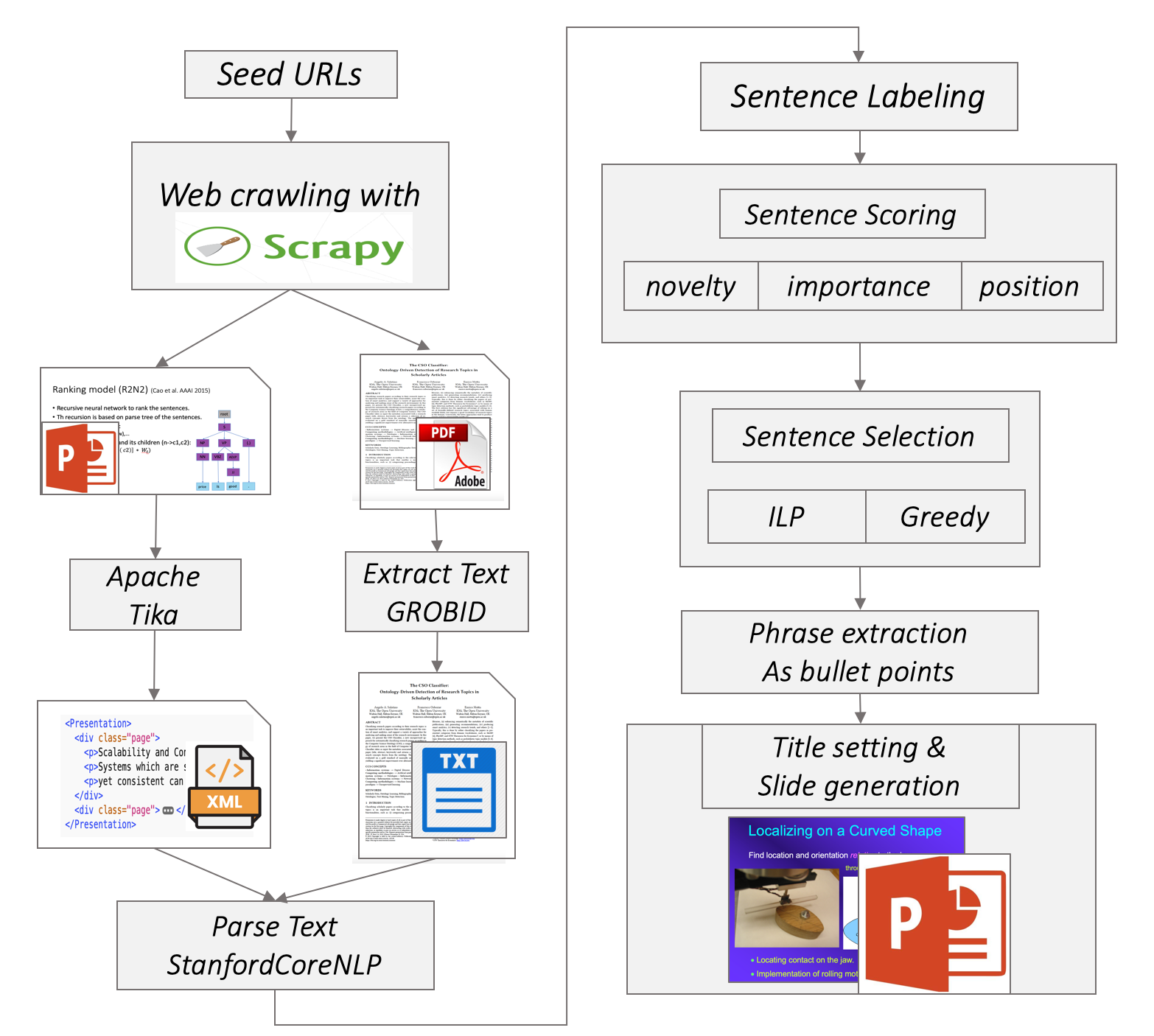}
\caption{Main components of the model for summarizing the paper and building the slides.}
\label{framework}
\end{figure}

Our contribution is threefold.
\begin{itemize}
    \item Propose a system that utilizes sentences with high rankings for generating presentation slides for research papers and is used as a starting point in the slide generation process.
    \item 
    Create and provide PS5K, a corpus of 5000 paper-slide pairs in the field of \textbf{computer} and \textbf{information science}. To the best of our knowledge, this is the largest paper-slide dataset and can be used for training and evaluating slide generation models.  
    \item 
    Propose a novel method to rank sentences within a sentence window, which improved an existing state-of-the-art text-summarization method by a significant margin. 
\end{itemize}

\section{Related Work}

Summarizing scholarly articles in presentation slides is different from standard text summarization~\cite{xiao2019emnlp}, which focuses on generating a paragraph of free text summary out of a longer document. Automatic slide generation can be done by first extracting salient sentences in a hierarchical order and grouping them into slides that are sequentially aligned with the original paper.


PPSGen \citep{hu2014ppsgen} was a framework that automatically generated presentation slides from scientific papers. They applied Support Vector Regressor and Integer Linear Programming (ILP) to rank and select important sentences. \citet{Wang2017PhraseBasedPS} generate slides by extracting phrases from papers and learning the hierarchical relationship between pairs of phrases to build the structure of bullet points. Their model is trained on a small set of 175 paper-slide pairs. The slideSeer \citep{kan2007slideseer} project crawled more than 10,000 paper-slide pairs using the Google APIs to search for the slide of papers using their title as a search query. The full set of data is not publicly available (only 20 pairs are available). Compared with previous works, our model is trained and tested on a relatively large set of 5000 paper-slide pairs and the dataset will be publicly available for future works. There had been some work on the alignment of presentations slides to the article sections \cite{hayama2005alignment, kan2007slideseer,beamer2009investigating}.

SummaRuNNer \citep{nallapati2017summarunner} is a neural extractive summarizer that treats the summarization task as a sequence labeling problem. 
SummaRuNNer was evaluated on CNN/Daily Mail corpus, which contains news articles that are shorter than research papers.
We improve upon the SummaRuNNer model for the summarization of scientific papers. 

\section{Data}

Producing a large dataset for summarization of scientific documents is challenging and  requires domain experts to make the summaries. The latest CL-Scisumm 2018 summarization task contains \textbf{only 40 NLP papers} with human-annotated reference summaries. Recently, ScisummNet \cite{yasunaga2019scisummnet} expanded the CL-Scisumm to 1000 scientific articles. TalkSum \cite{lev2019talksumm} summarizes scientific articles based on the transcripts of the presentation talks at conferences.

Using presentation slides made by the authors is promising for the training of deep neural summarization models as more conferences are providing slides with papers.


We crawled more than 5,000 paper-slide pairs from a manually curated list of websites, e.g., usenix.org and aclweb.org.  
GROBID \cite{GROBID} is used to get metadata and the body of the text from scientific papers in PDF format. Presentations are transformed form PDF or PPT format to XML by Apache Tika\footnote{https://tika.apache.org/}. The Tika XML files are divided into $pages$ and the text is extracted using Optical Character Recognition (OCR) tools.
Most venues of papers in our dataset are in computational linguistics, system, and system security. 
In our dataset, there are on average 35 pages of slide per presentation and 8 lines of text per slide page. The majority (75\%) of papers are published between 2013 and 2019. 
We used this dataset (called PS5K) to train summarization models to identify important parts of the input document at the sentence level.

\section{Method}
Generating slides requires identifying important sentences of the input scientific article and consists of three main steps. The first is to label salient sentences in the paper that are literally similar to corresponding slides. The second is to train the model to rank sentences and the final step selects salient sentences based on the predicted scores, size of the summary and the length of the sentences.
Afterwards, frequent noun phrases are extracted from the selected sentences to shape the hierarchical structure of the bullet points. The architecture of our model is shown in Figure \ref{framework}.

\subsection{Sentence Labeling}
The text in manually generated slides may not be directly extracted from the original paper. Instead, text can be truncated, summarized, or rephrased. Therefore, we need to generate extractive labels for sentences of the input document. The sentence labeling process attempts to identify salient sentences that are semantically similar to the corresponding slides. This generates an \emph{extractive} summary, which will be used as the ground truth for training and evaluation. 

The problem is formalized below:

A research paper can be represented as a sequence of $n$ sentences $D=\{s_1, s_2, ... s_n\}$, each having a label $y_i \in\{0,1\}$, the system predicts $p(y_i=1)$, probability of including sentence $i$ to the summary.

SummaRuNNer treats the summarization task as a sequence labeling problem, if adding the sentence to the summary improves the ROUGE score, the sentence is labeled with 1, otherwise it is labeled with 0. This method is suitable for news articles such as CNN/DailyMail \cite{nallapati2016abstractive} where the first couple of sentences in articles usually cover the main content. Scholarly papers usually contain a hierarchical structure of sections. Each section should have its own summary as a part of the summary of the entire paper. Therefore, the labeling process should be adapted to distribute positive labels across all sections of the paper. However, accurately parsing sections of open domain scholarly papers is non-trivial. Therefore, we propose a windowed labeling approach, in which ranking is performed only within a series of non-overlapping text windows, each of which contains $w$ consecutive sentences. A sentence is labeled as 1 if adding the current sentence increases the ROUGE-1 index. 
The best window size is determined empirically by trying different widow sizes and calculating the ROUGE score between selected sentences and the presentation slides. Section \ref{experiment} elaborates on the experiments performed to select the best window size.

\subsection{Sentence and Document Embedding} \label{ranker}


The ranking of sentences depends on their salience, novelty, and content similarity to the ground truth. To quantify these characteristics, a document is represented into a vector. We explore two methods to build the embedding for the whole document. 

\paragraph{Simple Document Embedding}
\newcommand{\cev}[1]{\reflectbox{\ensuremath{\vec{\reflectbox{\ensuremath{#1}}}}}}
A simple document embedding can be obtained by calculating the average of sentence encodings generated by a Bi-directional Long Short-Term Memory (BiLSTM) \citep{hochreiter1997long}. A sentence $s_i$ can be encoded as 
$ E_{s_i} = [\vec{h_i},\cev{h_i}] $ in which $E_{s_i}$ is a concatenation of forward ($\vec{h_i}$) and backward ($\cev{h_i}$) hidden states of the last token in sentence $s_i$. The embedding for document $D$ with $n$ sentences is the average of all sentence embeddings: 
\begin{equation}
    E_D = ReLU( W\times \frac{1}{n} \sum_{i=1}^n E_{s_i} + b)
\end{equation}

in which $ReLU$ is the activation function, $W$ and $b$ are parameters to be learned.

\paragraph{Hierarchical Self Attention Document Embedding}
This model embeds a document by applying the attention mechanism at both word and sentence levels \cite{al2018hierarchical, yang2016hierarchical}.

\emph{Sentence embeddings} are obtained by encoding word-level tokens of a sentence using BiLSTM and then aggregating hidden layers using an attention mechanism. 
Formally, considering a sentence $s_i$ with $m$ words,
the sentence encoding $h_{s_i}$ is obtained as a concatenation of all $m$ hidden states of word-level tokens $ (h_{s_i} = [h_1, h_2, ..., h_m]) $ where $h_{s_i} \in \mathbb{R}^{m \times 2d}$ and $d$ is the embedding dimension for each word.
The attention weights are:
\begin{equation}
    a_{\rm word} = softmax \left(W_{\rm attn} \times h_{s_i}^T\right)
\end{equation}
where $ W_{\rm attn} \in\mathbb{R}^{k\times2d}$ is the model matrix to be learned. Then $a_{\rm word} \in\mathbb{R}^ {k \times m}$ and the embedding for sentence $s_i$ is:
\begin{equation}
E_{s_i} = \mathop{average}_{k} \left( a_{word} \times h_{s_i} \right)
\end{equation}
where $E_{s_i} \in\mathbb{R}^{ 1\times 2d}$ and $k$ is the attention dimension which is set to 100 in our experiments.

\emph{Document embeddings} ($E_D$) are generated using sentence embeddings ($E_{s_i}$) built in the previous step. A similar attention layer is applied on top of sentence embeddings to build the document embedding. The sentence level attention works as the weights to emphasize important sentences in document embedding.



\subsection{Sentence Ranking }

The rank of a sentence depends on its position in the paper, salience, and novelty with respect to the previously selected sentences, calculated below: 
\begin{equation}
\begin{gathered}
    pos = position \times W_{pos} \\
    content = E_{s_i} \times W_{content} \\
    salience = E_D \times W_{salience} \times E_{s_i}^T \\
    novelty = summary_i \times W_{novelty} \times E_{s_i}^T \\
    p(y_i =1) = \sigma(pos + content + novelty + \\ salience) \\
\end{gathered}
\end{equation}

where $W_{pos} \in\mathbb{R}^{ 2d \times 1} $,where $W_{content} \in\mathbb{R}^{ 2d \times 1} $ $W_{salience} \in\mathbb{R}^{ 2d \times 2d}$, and $W_{novelty} \in\mathbb{R}^{ 2d \times 2d}$ are parameters to be learned. The $position$ is the position of the sentence in the document specified by a Embedding lookup function, $\sigma$ is the sigmoid activation function, and $pos$ is its positional embedding. The $salience$ estimates the importance of a sentence. The $novelty$ represents the novelty of a sentence with respect to the current summery. The summary embedding is the weighted sum of the previous sentences added to summary until sentence $i$:
\begin{equation}
    summary_i = \sum_{j=0}^{i-1} p(y_i=1) \times E_{s_i}
\end{equation}

\begin{figure}[htbp]
\includegraphics[width=0.48\textwidth, height=4cm]{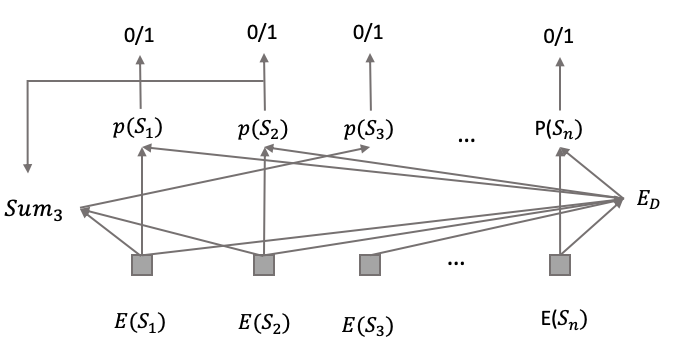}
\caption{Score prediction for sentence 3 depends on document embedding ($E_D$), sentence embedding, the embedding of the summary built until step 3 ($Sum_3$), and position of the sentence which is 3. The summary is the weighted sum of the embeddings of the first and second sentences. }
\label{sumarunner}
\end{figure}

The higher chance of adding the sentence to the summary gives it a bigger portion in the summary embedding. 
Figure \ref{sumarunner} shows the architecture for predicting the score for the third sentence in a document.

With windowed labeling, the positive labels are sparse. To deal with the imbalanced positive labels, the following weighted cross-entropy loss is adopted. The setting of $w_1=-85$ and $w_2=-2$ results in the highest ROUGE score. 
\begin{equation}
\begin{gathered}
    loss = - \sum_{i=0}^{n} w_1 y_i \times\log \left(p(y_i=1)\right) \\
    + w_2 (1-y_i) \times\log \left(1-p(y_i=1)\right)
\end{gathered}
\end{equation} 
\begin{table*}
  \caption{ROUGE scores for different models. Oracle and TextRank are unsupervised and do not need training.  $T_{\rm tr}$ standards for training time in hours based on Nvidia GTX 2080 Ti GPU. SRNN stands for SummaRuNNer.}
  \label{ROUGE_score}
  \centering
  \begin{tabular}{l|c|c|c|c}
  \toprule
     \multicolumn{1}{c|}{ Models} &  ROUGE-1 &   ROUGE-2&   ROUGE-L & $T_{\rm tr}$ \\
    \midrule
    Oracle (window=10) & 57.12 & 16.53 & 27.62 & - \\ 
    Sefid et al. \cite{sefid2019automatic} & 36.33 & 8.73 & 17.02 & - \\
    TextRank \cite{DBLP:journals/corr/BarriosLAW16} & 38.87 & 9.28 & 19.75& - \\
    SRNN+ILP & 45.12 & 11.65 & 22.96 & 18\\
    SRNN+greedy & 45.04 & 11.67 & 23.03  & 18\\
    Attn+windowed SRNN+ILP & 47.49 & 11.67 & 22.89 & 38\\
    Attn+windowed SRNN+greedy & 47.56 & 11.68 & 23.30 & 38\\
    windowed SRNN+ILP & \textbf{48.29} & 12.00 & \textbf{23.80} & 18\\ 
    windowed SRNN+greedy & 48.28 & \textbf{12.02} & 22.14 & 18\\ 
   \bottomrule
  \end{tabular}
\end{table*}

\subsection{Sentence Selection}
 To select the sentences for the slide we tried 1) the greedy approach that sequentially adds sentences with highest scores until the maximum limit is hit and 2) the ILP method that selects the sentences by optimizing the following function using IBM CPLEX Optimizer \footnote{https://www.ibm.com/products/ilog-cplex-optimization-studio}.
\begin{equation}
\begin{gathered}
    \max \sum_{i\in N_s} l_ix_i \times p(y_i=1)\\
    \sum_{i} l_ix_i < maxLen,\quad \forall i,\ x_i \in \{0,1\}
\end{gathered}
\end{equation}

where $p(y_i=1)$ is the score of the sentence predicted by the model, $x_i$ is a binary variable showing whether sentence $i$ is selected for the summary or not, $l_i$ is the length of sentence $i$ and penalizes short sentences, and $maxLen$ is the maximum length of the summary. 
\subsection{Slide Generation} 
A typical presentation slide includes a limited number of bullet points as the first-level, which are usually phrases or shortened sentences. Some slides may contain second-level bullet points for further breakdowns. Table~\ref{bullet_stats} shows that less than 8\% of the content of the presentations in the ground truth corpus is covered in third-level bullets. We generate slides containing up to 2 bullet levels.
Table~\ref{bullet_stats} also shows that a slide title on average contains 4 words and either Level 1 or Level 2 bullets contains on average 8 words.  Each slide consists of on average 36 words in 5 bullets and each level-1 bullet includes 2 second-level bullets.  

Sentences selected are treated as the second-level bullets. The first-level bullets are the noun phrases extracted from the sentences.  
Noun phrases are removed if they contain more than 10 words or just 1 word. Noun phrases with a document frequency greater than 10 are excluded (e.g. ``the model''). The section, which the first sentence of a slide is in, is found and its heading is  used as the slide title. The heading is truncated to the first 5 tokens. We limit a maximum of 4 sentences per slide. If a topic has more than 4 related sentences, the slide is split into two distinct ones. 

\begin{table}
  \caption{Bullet points statistics.}
  \label{bullet_stats}
  \centering 
  
  \begin{tabular}{ccc}
    \toprule
    Bullet-Point & Fraction &  Avg Word Count\\
    \midrule
    Title & - & 3.7 \\
    Level 1 & 56.5\% & 7.38 \\ 
    Level 2 & 35.5\% & 7.22 \\
    Level 3 & 7.9\% & 6.7 \\
    \bottomrule
  \end{tabular}
\end{table}

\section{Experiments and Results} \label{experiment}

We estimated the parameters of our model on PS5K. We split the dataset into training, validation, and testing set, each consisting of 4500, 250, and 250 pairs, respectively. We experimented with different window sizes 
and found that a window size of $w=10$ gives the best ROUGE-1 recall (Table \ref{window}) and is adapted for our model. 

\begin{table}
  \caption{ROUGE scores for oracle summaries generated with different window sizes.}
  \label{window}
  \centering
  \begin{tabular}{c c c c}
  \toprule
    Window Size &  R-1 &   R-2 &   R-L \\
    \midrule
    3 & 42.95 & 11.13 & 21.59 \\
    5 & 44.34 & 11.43 & 22.35 \\
    7 & 44.88 & 11.64 & 22.47 \\
    {\bf 10} &  {\bf 45.93} &  {\bf 12.00} &  {\bf 22.75} \\
    15 & 45.52 & 11.84 & 22.68 \\
    \bottomrule
  \end{tabular}
\end{table}

The Stanford CoreNLP \cite{manning-EtAl:2014:P14-5} is used to tokenize and lemmatize sentences to the constituent tokens and to extract noun phrases. GloVe \cite{pennington2014glove} 50-dimensional vectors are used to initialize the word embeddings.
With the AdaDelta optimizer and a learning rate of 0.1, we trained for 50 epochs. The sentences are truncated or padded to have 50 tokens (only 8\% sentences consist of more than 50 tokens). Similarly, we adopt a fixed document size of 500 sentences (only 3.5\% of documents in our dataset have more than 500 sentences).
We used the standard ROUGE score \cite{lin2004ROUGE} to evaluate the summaries. 
The ROUGE scores for summaries are tabulated in Table \ref{ROUGE_score}. The summary size can not exceed 20\% of the size of the input document in words. 
TextRank \cite{mihalcea2004textrank} is a graph based summarizer that applies the Google PageRank \cite{page1999pagerank} algorithm to rank the sentences. Sefid et al. \cite{sefid2019automatic} rank the sentences by combining surface features, semantic and contextual embeddings.
The windowed SummaRuNNer+ILP model outperforms the base SummaRuNNer by at least 3 points in ROUGE-1 recall. Adding attention layer to the model does not improve the ROUGE score while it increases the training time considerably as there are more parameters to be trained.

 \section{Conclusion}  We create and make available PS5K, which is a large slide-paper dataset consisting of 5,000 scientific articles and corresponding manually made slides. This dataset can be used for scientific document summarization and slide generation. 
We used state of the art extractive summarization methods to summarize scientific articles. Our results show that distributing the positive labels across all sections of a scientific paper, in contrast to summarization methods for news articles, considerably improves performance.
The code is available
\href{https://github.com/atharsefid/Extractive_Research_Slide_Generation_Using_Windowed_Labeling_Ranking}{here}.

\bibliography{anthology,custom}
\bibliographystyle{acl_natbib}

\end{document}